# Ethical AI in the Healthcare Sector: Investigating Key Drivers of Adoption through the Multi-Dimensional Ethical AI Adoption Model (MEAAM)

## Prathamesh Muzumdar [a*], Apoorva Muley [b], Kuldeep Singh [c] and Sumanth Cheemalapati [d]

[a] *University of Texas at Arlington, United States.*
[b] *People's University, Madhya Pradesh, India.*
[c] *Arkansas Tech University, Arkansas, United States.*
[d] *Dakota State University, United States.*

*Original Research Article*

## ABSTRACT

The adoption of Artificial Intelligence (AI) in the healthcare service industry presents numerous ethical challenges, yet current frameworks often fail to offer a comprehensive, empirical understanding of the multidimensional factors influencing ethical AI integration. Addressing this critical research gap, this study introduces the Multi-Dimensional Ethical AI Adoption Model (MEAAM), a novel theoretical framework that categorizes 13 critical ethical variables across four

---

foundational dimensions of Ethical AI: Fair AI, Responsible AI, Explainable AI, and Sustainable AI. These dimensions are further analyzed through three core ethical lenses: epistemic concerns (related to knowledge, transparency, and system trustworthiness), normative concerns (focused on justice, autonomy, dignity, and moral obligations), and overarching concerns (highlighting global, systemic, and long-term ethical implications). This study adopts a quantitative, cross-sectional research design using survey data collected from healthcare professionals and analyzed via Partial Least Squares Structural Equation Modeling (PLS-SEM). Employing PLS-SEM, this study empirically investigates the influence of these ethical constructs on two outcomes—Operational AI Adoption and Systemic AI Adoption. Results indicate that normative concerns most significantly drive operational adoption decisions, while overarching concerns predominantly shape systemic adoption strategies and governance frameworks. Epistemic concerns play a facilitative role, enhancing the impact of ethical design principles on trust and transparency in AI systems. By validating the MEAAM framework, this research advances a holistic, actionable approach to ethical AI adoption in healthcare and provides critical insights for policymakers, technologists, and healthcare administrators striving to implement ethically grounded AI solutions.

## 1. INTRODUCTION

Ethical Artificial Intelligence (AI) refers to the design, development, deployment, and governance of AI systems in ways that uphold moral principles such as justice, autonomy, accountability, transparency, and human well-being. It involves aligning AI technologies with societal values, ensuring responsible decision-making, and mitigating unintended harms across various applications. As AI continues to permeate multiple industries, ethical AI has emerged as a critical consideration in shaping trustworthy, fair, and socially beneficial systems (Reddy et al., 2020; Paulus & Kent, 2020). The healthcare sector has become a prominent domain for AI integration due to its immense potential to improve clinical outcomes, streamline diagnostics, enhance operational efficiency, and expand access to care (Khan et al., 2022; Prosperi et al., 2020). AI-driven solutions such as predictive analytics, medical imaging interpretation, robotic surgeries, and personalized treatment planning are increasingly being embedded into healthcare delivery systems (Beil et al., 2019, Tran et al., 2019). However, the integration of AI into healthcare introduces ethical complexities related to data privacy, decision accountability, patient safety, and equity (Séroussi et al., 2020). These concerns make it imperative to examine how ethical considerations influence the adoption and sustained use of AI technologies in healthcare institutions.

Ethical AI in healthcare can be understood through a typology of key paradigms. Fair AI emphasizes governance and social justice;

Human-Centered AI focuses on augmenting, not replacing, human capabilities; Responsible AI ensures accountability, safety, and long-term societal well-being; Privacy-Preserving AI safeguards user data and ensures informed consent through techniques such as federated learning and differential privacy; Transparent AI prioritizes openness and interpretability; Explainable AI aims to demystify complex algorithms; Beneficent AI is oriented toward doing good and promoting human flourishing; Inclusive AI addresses accessibility and multicultural inclusiveness; and Sustainable AI emphasizes environmental and social sustainability (Paulus & Kent, 2020). To empirically investigate how these ethical paradigms influence AI adoption, we operationalize a set of key variables. These include Justice and Fairness, Freedom and Autonomy, Privacy, Transparency, Patient Safety, Cyber Security, Trust, Beneficence, Responsibility, Solidarity, Sustainability, Dignity, and Conflict. Each variable aligns with one or more types of Ethical AI (Ahmad et al., 2020). For example, "Justice and Fairness" maps to Fair AI, "Privacy" to Privacy-Preserving AI and Responsible AI, and "Trust" to Human-Centered and Responsible AI.

Furthermore, these variables also represent distinct ethical concerns that are essential to understanding the broader impact of AI in healthcare (Fiske et al., 2019). Epistemic concerns focus on how AI generates and communicates knowledge, and are reflected in variables such as Transparency, Trust, and Cyber Security (He et al., 2019). Normative concerns emphasize rights, duties, and moral obligations, and are represented through Justice and Fairness, Freedom and Autonomy, Privacy, Patient Safety, Beneficence, Responsibility, and Dignity (Rajkomar et al., 2018). Overarching concerns relate to systemic, societal, and long-term issues, captured through variables like Solidarity, Sustainability, and Conflict (Balthazar et al., 2018). The dependent construct in this study is AI adoption, conceptualized as a five-stage process: Aware, Active, Operational, Systemic, and Transformational (Bæ røe et al., 2020). These stages reflect an organization's maturity and commitment in integrating AI into healthcare systems—from initial awareness to full-scale, strategic transformation. However, rather than treating adoption as a monolithic outcome, this study focuses on two critical stages: the Operational stage, where AI becomes embedded in workflows, and the Systemic stage, where AI is integrated holistically across the healthcare institution.

Despite growing attention to AI ethics and adoption, a notable research gap remains in understanding the causal relationships between ethical considerations and AI adoption outcomes (Vellido, 2019). Particularly, there is limited empirical analysis exploring how variables representing different types of Ethical AI influence progress into the operational and systemic stages of adoption. To address this gap, this study introduces the Multi-Dimensional Ethical AI Adoption Model (MEAAM)—a conceptual framework developed to empirically analyze the interplay between ethical AI dimensions and distinct adoption stages in the healthcare sector (Starke et al., 2021). By investigating these relationships, this study aims to bridge the theoretical and empirical divide, providing evidence-based insights for policymakers, developers, and healthcare leaders striving to implement AI ethically and effectively.

## 2. LITERATURE REVIEW AND THEORETICAL MECHANISM

### 2.1 Relevant Studies on Ethical AI and AI Adoption

The intersection of ethical AI and adoption in healthcare has been the subject of increasing academic interest in recent years. Scholars have examined the potential of AI to enhance healthcare delivery, improve diagnostic accuracy, and optimize resource management (Yew, 2021). However, a parallel line of inquiry has highlighted the ethical challenges posed by AI systems, such as algorithmic bias, lack of transparency, and threats to privacy and autonomy (Wiens et al., 2019). These challenges are particularly critical in healthcare, where decisions can have profound impacts on human life and well-being (Geis et al., 2019). Previous studies have largely focused on ethical principles in a conceptual manner or have assessed adoption readiness based on technological or organizational factors. Few studies have attempted to integrate these two dimensions—ethical AI and the adoption process—into a single analytical framework, especially through empirical methods.

### 2.2 Types of Ethical AI and Variables

Recent literature has identified several typologies of ethical AI, each focusing on specific value-driven goals and principles. Fair AI emphasizes equity, bias mitigation, and societal impact, while Human-Centered AI aims to enhance rather than replace human capacities (Morley & Floridi, 2019a). Responsible AI is closely tied to governance, accountability, and safety. Privacy-Preserving AI leverages advanced techniques like federated learning and differential privacy to protect user autonomy and data. Transparent and Explainable AI models seek to make decision processes understandable and interpretable. Other ethical frameworks include Beneficent AI, which promotes well-being; Inclusive AI, which addresses access and representation; and Sustainable AI, which emphasizes environmental and long-term considerations (Currie et al., 2020). Each type of Ethical AI is operationalized through specific variables, such as Justice and Fairness, Privacy, Trust, Responsibility, and Dignity (Panesar & Panesar, 2020). These variables serve as measurable indicators that can help assess the ethical orientation of AI systems in practice.

### 2.3 Types of Concerns and Outcomes

In addition to categorizing AI through ethical types, literature also distinguishes the concerns these variables address. Epistemic concerns relate to how AI systems produce, justify, and communicate decisions, and include issues such as Transparency, Trust, and Cyber Security (Morley & Floridi, 2019b). Normative concerns focus on rights, values, and moral obligations, and are reflected in variables like Justice and Fairness, Privacy, Patient Safety, and Responsibility (Lekadir et al., 2022). Overarching concerns refer to the broader societal and long-term implications of AI deployment and include issues such as Solidarity, Sustainability, and Conflict (Nikitas et al., 2020). This categorization helps in understanding not only what values are at stake, but also the levels at which ethical tensions operate—individual, institutional, or societal.

## 2.4 Levels of AI Adoption

The literature on AI adoption presents a multi-stage process, where organizations progress through increasing levels of engagement with AI technologies. These stages include Aware (recognition of AI's potential), Active (initial experimentation or pilot projects), Operational (AI integrated into specific functions), Systemic (AI aligned with broader organizational strategy), and Transformational (AI driving innovation and restructuring) (D'antonoli, 2020; Cohen et al., 2014). Most research on AI adoption focuses on technological readiness, leadership commitment, or organizational capabilities, with relatively little emphasis on how ethical factors influence progression through these stages (Kaissis et al., 2020). Furthermore, while qualitative insights are available on ethical resistance or public perception, there is a lack of empirical models that connect ethical dimensions directly with adoption outcomes.

This study addresses a key research gap by empirically exploring the causal relationship between variables representing different types of Ethical AI and two specific stages of AI adoption—Operational and Systemic. These stages are critical milestones in the institutionalization of AI, where ethical considerations are no longer peripheral but central to sustained and scalable integration. By linking ethical values to adoption levels, this study contributes to a more holistic understanding of how ethical readiness may serve as both a facilitator and a gatekeeper in the healthcare sector's AI transformation.

## 2.5 Research Gap

The existing literature on ethical AI and AI adoption in healthcare offers valuable conceptual frameworks and qualitative insights into the importance of ethics in shaping trust, usability, and social acceptance of AI systems. However, as established in the previous section, much of this research has treated AI adoption as a singular or generalized outcome within behavioral or organizational studies, without differentiating between the stages at which adoption occurs. This generalized treatment overlooks the nuanced ways in which ethical considerations might influence different levels of adoption maturity—particularly the operational and systemic stages that are critical for the sustainable integration of AI in healthcare institutions (Andorno, 2004). By dissecting AI adoption into distinct stages, this study introduces a novel empirical pathway to understanding how ethical variables drive progression through key adoption milestones. The operational stage represents the point at which AI becomes embedded in specific healthcare workflows, requiring a balance between ethical safeguards and functional efficiency (Cookson, 2018). The systemic stage, on the other hand, reflects a broader organizational commitment where AI aligns with institutional goals and ethical values are institutionalized across decision-making and strategic planning. These stages demand different levels of preparedness, governance, and ethical alignment, which cannot be fully captured through an undifferentiated view of adoption (Cruz Rivera et al., 2020).

**Table 1. Types of ethical AI**

| Type of Ethical AI | Definition |
|---|---|
| 1. **Fair AI** | Governance, accountability, safety, long-term societal impact<br>Ensures that AI systems are built and used responsibly. |
| 2. **Human-Centered AI** | Enhancing human capabilities, not replacing them<br>AI should work with and for people, respecting human values and dignity. |
| 3. **Responsible AI** | Governance, accountability, safety, long-term societal impact<br>Ensures that AI systems are built and used responsibly. |
| 4. **Privacy-Preserving AI** | Data protection, user autonomy, informed consent<br>Ensures personal data is securely handled and anonymized.<br>Includes methods like:1. Federated learning 2. Differential privacy |
| 5. **Transparent AI** | Explainability, interpretability, openness<br>AI decisions must be understandable and traceable by humans. |
| 6. **Explainable AI** | Making AI models and decisions interpretable by humans<br>A technical and ethical effort to open the "black box" of complex models like deep learning. |
| 7. **Beneficent AI** | Doing good, promoting human flourishing<br>AI should contribute positively to well-being, education, healthcare, etc.<br>Avoids unintended harm (non-maleficence) |
| 8. **Inclusive AI** | Accessibility, multicultural perspectives, underserved groups<br>Ensures marginalized or diverse populations are represented and served by AI. |
| 9. **Sustainable AI** | Environmental and social sustainability<br>Designs AI systems that reduce energy consumption and support long-term ecological balance. |

**Table 2. Description of variables and AI type**

| Variable | Primary Type of Ethical AI | Explanation |
|---|---|---|
| 1. Justice & Fairness | Fair AI | Core principle focused on equality, bias mitigation, and social justice. |
| 2. Freedom & Autonomy | Human-Centered AI / Responsible AI | Ensures individuals retain control over their data and decisions. |
| 3. Privacy | Privacy-Preserving AI / Responsible AI | Protects personal data and respects informed consent. |
| 4. Transparency | Transparent AI / Explainable AI | Demands interpretability and openness in AI processes. |
| 5. Patient Safety | Responsible AI | Emphasizes harm prevention and reliability, especially in health contexts. |
| 6. Cyber Security | Responsible AI / Privacy-Preserving AI | Protects systems and data from unauthorized access or manipulation. |
| 7. Trust | Responsible AI / Human-Centered AI | Built through consistent, safe, and understandable AI behavior. |
| 8. Beneficence | Beneficent AI / Responsible AI | Encourages AI that improves human well-being and minimizes harm. |
| 9. Responsibility | Responsible AI | Ensures accountability in design, deployment, and impact. |
| 10. Solidarity | Inclusive AI / Fair AI | Encourages AI systems that support collective good and equity. |
| 11. Sustainability | Sustainable AI | Promotes environmental and societal sustainability in AI development. |
| 12. Dignity | Human-Centered AI / Responsible AI | Respects human value and prevents dehumanization by AI systems. |
| 13. Conflict | Responsible AI | Addresses ethical trade-offs and stakeholder tensions responsibly. |

**Table 3. Types of AI concerns & variables**

| Type of Concern | Variables | Definition |
|---|---|---|
| 1. Epistemic Concerns | Transparency<br>Trust<br>Cyber Security | Ethical issues related to how AI systems generate, justify, and communicate knowledge and decisions |
| 2. Normative Concerns | Justice & Fairness<br>Freedom & Autonomy<br>Privacy<br>Patient Safety<br>Beneficence<br>Responsibility<br>Dignity | Ethical questions about what values, rights, and duties should guide the design and use of AI |
| 3. Overarching Concerns | Solidarity<br>Sustainability<br>Conflict | Broad, systemic, and long-term ethical implications of AI affecting societies, institutions, and global governance. |

**Table 4. Five stages of AI adoption**

| Five Stages of AI adoption | Definition |
|---|---|
| Aware | Organizations become conscious of AI's potential, its use cases, and emerging trends. At this stage, interest is building, but there is limited or no experimentation. |
| Active | Initial pilot projects or small-scale experiments are launched. Teams explore how AI could solve specific problems but results are not yet fully integrated into workflows. |
| Operational | AI solutions are embedded in core processes. The organization begins to standardize tools and platforms for consistent use across departments. |
| Systemic | AI is strategically aligned with business or policy goals. Governance, ethics, and performance monitoring are formalized. Data infrastructure and talent are scaled. |
| Transformational | AI becomes a key driver of innovation, cultural change, and competitive advantage. It reshapes decision-making, workflows, and creates new value propositions. Often includes continuous learning and responsible innovation. |

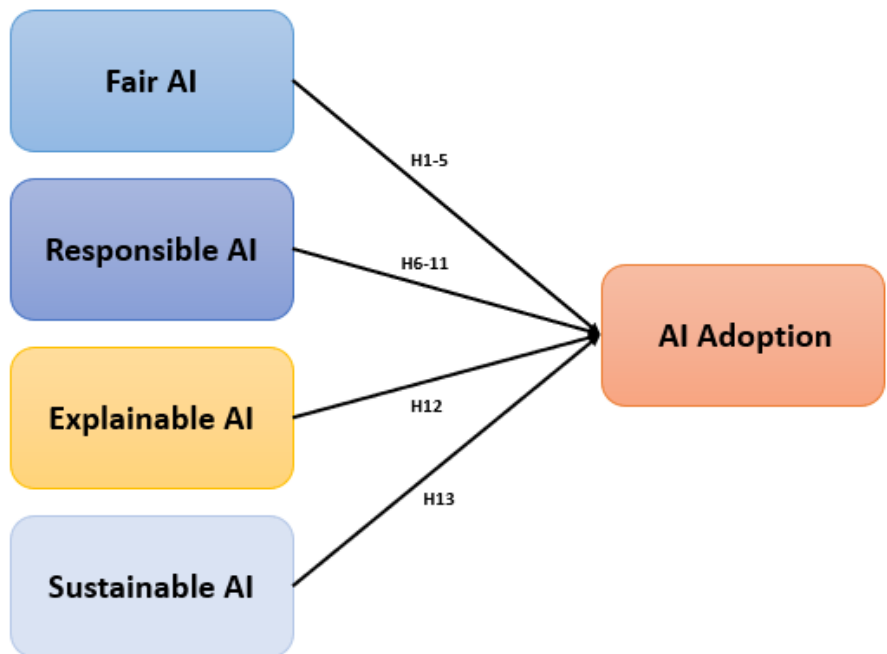

**Fig. 1. Conceptual model**

Previous studies have not sufficiently addressed how specific ethical variables—such as justice, autonomy, transparency, and beneficence—map onto these distinct adoption levels This gap limits our understanding of how ethical AI frameworks translate into real-world institutional change. Moreover, the lack of empirical investigation into causal relationships between ethical considerations and adoption levels constrains both academic theory-building and practical decision-making. Addressing this gap offers dual benefits. From an academic perspective, it allows for the development of a more granular theoretical model of AI adoption that incorporates ethical readiness as a dynamic driver of institutional change. From an industry perspective, understanding these relationships can inform more effective resource allocation, allowing healthcare organizations to budget strategically and design ethically aligned operational processes. This dual lens of theoretical advancement and practical relevance forms the foundation of this study's contribution to the emerging discourse on ethical AI in healthcare.

To fill this critical research void, the present study introduces the Multi-Dimensional Ethical AI Adoption Model (MEAAM)—a conceptual and empirical framework that links specific types of ethical AI (such as Fair AI, Human-Centered AI, Privacy-Preserving AI, Responsible AI, and others) with two key stages of AI adoption: operational and systemic. The MEAAM model is designed to empirically test the causal relationships between ethical constructs and adoption maturity, using real-world data collected from diverse stakeholders within the healthcare ecosystem. By doing so, the model bridges the gap between theory and practice, offering scholars a refined analytical lens for understanding adoption processes, while simultaneously guiding practitioners toward ethically informed AI implementation strategies. MEAAM stands as both a theoretical innovation and a practical decision-making tool, positioning ethics not as an afterthought but as a central driver of successful and sustainable AI integration in healthcare systems.

## 2.6 Multi-Dimensional Ethical AI Adoption Model (MEAAM)

The Multi-Dimensional Ethical AI Adoption Model (MEAAM) is proposed as a conceptual and analytical framework to empirically investigate the complex relationship between ethical principles and the adoption of AI in the healthcare sector. Recognizing that AI integration in healthcare involves more than just technical capability, MEAAM captures the ethical plurality of AI by organizing thirteen distinct ethical drivers—such as justice & fairness, privacy, transparency, and trust—into a unified model. Each driver corresponds to a recognized type of Ethical AI and is categorized by its associated ethical concern, whether epistemic, normative, or overarching. The model then maps these independent ethical dimensions to two critical stages of AI adoption: the operational stage, where AI becomes embedded in daily healthcare processes, and the systemic stage, where AI transforms institutional workflows and policies. This multi-layered design enables a nuanced understanding of how ethical values shape and influence the trajectory of AI implementation within healthcare systems.

MEAAM's strength lies in its integrative and empirical approach. Unlike traditional models that treat AI adoption as a singular behavioral outcome, MEAAM disaggregates adoption into developmental stages, allowing researchers and practitioners to identify which ethical drivers are most influential at specific points in the adoption lifecycle. By applying Partial Least Squares Structural Equation Modeling (PLS-SEM), the model evaluates the causal impact of each ethical driver on the progression of AI adoption. From an academic standpoint, MEAAM contributes to the development of a structured theory of ethical AI implementation. From an industry perspective, it offers healthcare administrators and policymakers a practical tool for ethically grounded decision-making, resource allocation, and implementation strategy. As AI continues to transform healthcare delivery, MEAAM positions itself as a timely and relevant model for aligning innovation with human values.

## 2.7 Research Question and Hypothesis

This study investigates the influence of ethical dimensions of AI on two critical stages of AI adoption—operational and systemic—within the healthcare sector. As artificial intelligence becomes increasingly embedded in healthcare services, understanding how ethical considerations shape its adoption at different organizational levels becomes essential for both academic and practical advancements. Drawing from the theoretical foundations of Ethical AI, this study empirically evaluates how different ethical principles, conceptualized as independent variables, impact the progression of AI integration from routine operational processes to broader systemic transformation.

The central research question guiding this study is:

**"How do ethical dimensions derived from different types of Ethical AI—such as fairness, transparency, privacy, and responsibility—influence the progression from operational to systemic stages of AI adoption in the healthcare sector?"**

To explore this, thirteen independent variables representing ethical AI dimensions have been identified and matched with corresponding stages of AI adoption. The first set of hypotheses addresses how variables related to justice and fairness, freedom and autonomy, privacy, and transparency influence AI adoption. We hypothesize that greater emphasis on justice and fairness in AI development promotes both operational (H1a) and systemic (H1b) adoption by fostering equity and reducing bias. Similarly, prioritizing individual freedom and autonomy is expected to increase trust and acceptance of AI at operational (H2a) and systemic levels (H2b). Privacy-preserving AI practices, which ensure secure and informed use of personal data, are likely to support operational (H3a) and systemic (H3b) adoption. Finally, transparency in AI—making processes and outputs understandable—should be positively associated with both operational (H4a) and systemic (H4b) stages.

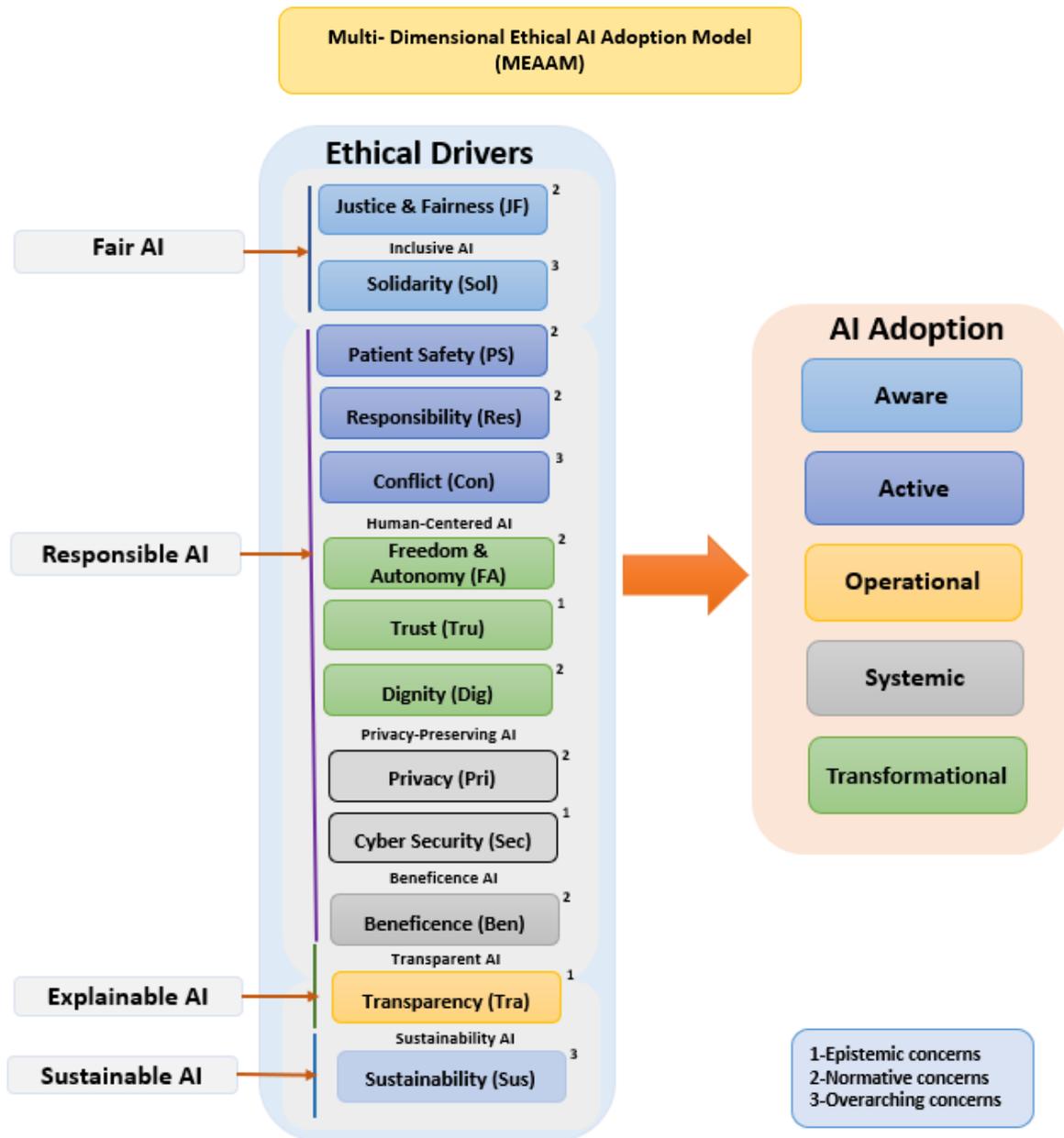

**Fig. 2. Multi-dimensional ethical AI adoption model (MEAAM)**

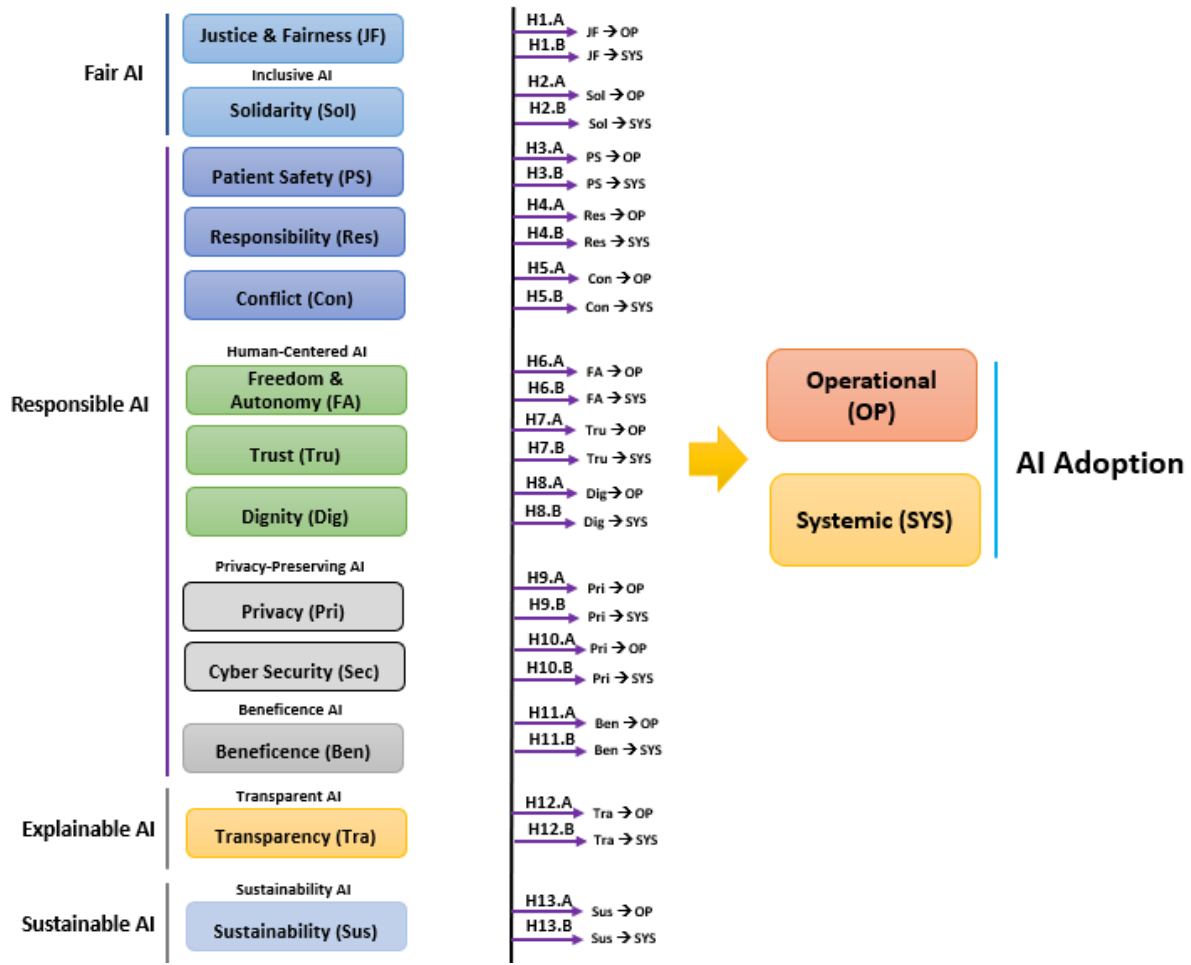

**Fig. 3. Conceptual design**

The second set of hypotheses explores the roles of patient safety, cybersecurity, trust, and beneficence in AI adoption. We expect that stronger attention to patient safety contributes positively to operational (H5a) and systemic (H5b) adoption by mitigating risks in sensitive environments. Robust cybersecurity practices are essential in healthcare and are hypothesized to support operational (H6a) and systemic (H6b) integration. Trust in AI systems, built through consistency and alignment with human values, is expected to positively influence operational (H7a) and systemic (H7b) adoption. Additionally, AI designed with a focus on human well-being and beneficence is hypothesized to drive both operational (H8a) and systemic (H8b) use.

The next set of hypotheses investigates responsibility, solidarity, sustainability, and dignity. When responsibility and accountability mechanisms are clearly embedded in AI systems, organizations are more likely to adopt AI at operational (H9a) and systemic (H9b) levels. AI that promotes solidarity—supporting equity and underserved communities—is also expected to enhance operational (H10a) and systemic (H10b) adoption. Similarly, sustainable AI practices, which reduce environmental and societal costs, are anticipated to positively impact operational (H11a) and systemic (H11b) adoption. Respect for human dignity, a key component of human-centered AI, is likely to increase the acceptance and embedding of AI tools in both operational (H12a) and systemic (H12b) stages.

**Table 5. List of hypothesis**

| Independent Variable (Ethical Dimension) | Primary Type of Ethical AI | Dependent Variable | Hypothesis |
|---|---|---|---|
| Justice & Fairness | Fair AI | Operational AI Adoption | **H1A**: Higher emphasis on justice and fairness is positively associated with operational AI adoption in healthcare. |
| | | Systemic AI Adoption | **H1B**: Higher emphasis on justice and fairness is positively associated with systemic AI adoption in healthcare. |
| Freedom & Autonomy | Human-Centered / Responsible AI | Operational AI Adoption | **H2A**: Emphasis on individual freedom and autonomy positively influences operational AI adoption in healthcare. |
| | | Systemic AI Adoption | **H2B**: Emphasis on individual freedom and autonomy positively influences systemic AI adoption in healthcare. |
| Privacy | Privacy-Preserving / Responsible AI | Operational AI Adoption | **H3A**: Strong privacy safeguards are positively associated with operational AI adoption in healthcare. |
| | | Systemic AI Adoption | **H3B**: Strong privacy safeguards are positively associated with systemic AI adoption in healthcare. |
| Transparency | Transparent / Explainable AI | Operational AI Adoption | **H4A**: Transparent AI practices are positively associated with operational AI adoption in healthcare. |
| | | Systemic AI Adoption | **H4B**: Transparent AI practices are positively associated with systemic AI adoption in healthcare. |
| Patient Safety | Responsible AI | Operational AI Adoption | **H5A**: Greater attention to patient safety is positively associated with operational AI adoption in healthcare. |
| | | Systemic AI Adoption | **H5B**: Greater attention to patient safety is positively associated with systemic AI adoption in healthcare. |
| Cyber Security | Responsible / Privacy-Preserving AI | Operational AI Adoption | **H6A**: Robust cybersecurity measures are positively associated with operational AI adoption in healthcare. |
| | | Systemic AI Adoption | **H6B**: Robust cybersecurity measures are positively associated with systemic AI adoption in healthcare. |
| Trust | Responsible / Human-Centered AI | Operational AI Adoption | **H7A**: Higher trust in AI systems is positively associated with operational AI adoption in healthcare. |
| | | Systemic AI Adoption | **H7B**: Higher trust in AI systems is positively associated with systemic AI adoption in healthcare. |
| Beneficence | Beneficent / Responsible AI | Operational AI Adoption | **H8A**: AI systems designed to promote well-being are positively associated with operational AI adoption in healthcare. |
| | | Systemic AI Adoption | **H8B**: AI systems designed to promote well-being are positively associated with systemic AI adoption in healthcare. |

| Independent Variable (Ethical Dimension) | Primary Type of Ethical AI | Dependent Variable | Hypothesis |
|---|---|---|---|
| Responsibility | Responsible AI | Operational AI Adoption | **H9A**: Clear accountability structures are positively associated with operational AI adoption in healthcare. |
| | | Systemic AI Adoption | **H9B**: Clear accountability structures are positively associated with systemic AI adoption in healthcare. |
| Solidarity | Inclusive / Fair AI | Operational AI Adoption | **H10A**: AI systems that promote solidarity and equity are positively associated with operational AI adoption in healthcare. |
| | | Systemic AI Adoption | **H10B**: AI systems that promote solidarity and equity are positively associated with systemic AI adoption in healthcare. |
| Sustainability | Sustainable AI | Operational AI Adoption | **H11A**: AI systems that support environmental and social sustainability are positively associated with operational AI adoption in healthcare. |
| | | Systemic AI Adoption | **H11B**: AI systems that support environmental and social sustainability are positively associated with systemic AI adoption in healthcare. |
| Dignity | Human-Centered / Responsible AI | Operational AI Adoption | **H12A**: Emphasis on protecting human dignity is positively associated with operational AI adoption in healthcare. |
| | | Systemic AI Adoption | **H12B**: Emphasis on protecting human dignity is positively associated with systemic AI adoption in healthcare. |
| Conflict | Responsible AI / Ethical Governance | Operational AI Adoption | **H13A**: Effective management of ethical conflicts is positively associated with operational AI adoption in healthcare. |
| | | Systemic AI Adoption | **H13B**: Effective management of ethical conflicts is positively associated with systemic AI adoption in healthcare. |

**Table 6. Stakeholder categories responding the multidimensional nature of AI implementation and its ethical implications**

| Stakeholder Group | Sample Size (n) | Percentage (%) | Rationale |
|---|---|---|---|
| 1. Healthcare IT Professionals | 52 | 14.0% | Key implementers and technical experts in AI tools. |
| 2. Senior Management / Executives | 47 | 12.6% | Strategic decision-makers influencing systemic AI integration. |
| 3. Senior Doctors / Clinicians | 61 | 16.4% | Clinical leaders overseeing AI-assisted medical processes. |
| 4. Junior Doctors / Residents | 58 | 15.6% | Frequent users of AI in diagnosis, monitoring, and decision support. |
| 5. Nursing Staff | 53 | 14.2% | Daily interaction with AI-based monitoring and patient care systems. |
| 6. Medical Technicians | 46 | 12.4% | Users of AI in diagnostic imaging, labs, and surgery-assistive tech. |
| 7. Patients / Patient Advocates | 55 | 14.8% | End-users impacted by AI decisions, offering a perspective on trust, transparency, and autonomy. |
| Total | 372 | 100% | |

Finally, the study hypothesizes that effective handling of ethical conflicts—especially in complex stakeholder environments—plays a critical role in successful AI implementation. Conflict resolution in AI ethics is thus expected to positively affect both operational (H13a) and systemic (H13b) stages of adoption. Together, these hypotheses offer a comprehensive empirical framework to examine how specific ethical AI variables can serve as drivers of AI adoption within the healthcare ecosystem.

## 2.8 Research Design

This study employs a quantitative, cross-sectional research design to empirically examine how ethical AI principles influence the progression of AI adoption within the healthcare sector. Drawing on established ethical AI frameworks, the research operationalizes thirteen independent variables representing diverse ethical dimensions—such as fairness, transparency, privacy, and responsibility—and investigates their impact on two specific stages of AI adoption: operational and systemic. Using a structured survey instrument, data will be collected from healthcare professionals, administrators, and technology officers engaged in AI-related decision-making. The design facilitates the testing of causal relationships through statistical methods such as multiple regression analysis, enabling a systematic evaluation of how ethical considerations affect the integration of AI into organizational workflows and strategic systems. By focusing on distinct levels of AI adoption, this approach aims to uncover targeted insights that can guide both theoretical modeling and practical implementation strategies in ethical AI governance.

## 2.9 Data Collection

To investigate the relationship between ethical AI dimensions and levels of AI adoption in the healthcare sector, this study utilized a structured survey-based data collection method. The target population comprised stakeholders who directly interact with or are impacted by AI-driven tools and processes within healthcare institutions. A total sample size of 372 respondents was selected through stratified random sampling to ensure representation across various professional roles and responsibilities related to AI usage. Data were collected over a timeframe extending from mid-2023 through early 2024. This approach supports a balanced understanding of both operational and systemic aspects of AI adoption by capturing the perceptions and experiences of both decision-makers and end users.

Respondents were recruited from healthcare organizations actively engaged in AI deployment, including hospitals, diagnostic centers, and telehealth providers. Participation criteria required that individuals either use AI systems in their daily operations or contribute to AI-related policy, governance, or technology development. To improve accessibility and response rate, an online survey was distributed through professional mailing lists, hospital affiliations, and healthcare forums. The survey was anonymous and included informed consent to ensure ethical compliance. Participants were grouped into seven key stakeholder categories to reflect the multidimensional nature of AI implementation and its ethical implications.

The final distribution of the 372 respondents is provided below Table 7.

This multi-stakeholder design ensures a holistic and grounded understanding of how ethical considerations shape AI adoption trajectories in real-world healthcare settings.

# 3. RESEARCH METHODS

## 3.1 PLS-SEM

Partial Least Squares Structural Equation Modeling (PLS-SEM) was chosen as the primary analytical method for this study due to its robust capacity to model complex relationships between latent constructs and to handle reflective measurement models (McDougall, 2019). PLS-SEM is particularly suitable when the objective is to explore and predict causal relationships rather than merely confirm existing theories. In the context of this study, which introduces the novel Multi-Dimensional Ethical AI Adoption Model (MEAAM), PLS-SEM offers a flexible framework to evaluate the influence of thirteen distinct ethical AI variables on two levels of AI adoption—operational and systemic. The technique is also highly tolerant of non-normal data distributions and works effectively with smaller to moderate sample sizes, which adds to its appeal in empirical health technology research involving diverse respondent groups like IT professionals, doctors, nurses, and patients. The data were analyzed using SmartPLS 4.0 software, which facilitated the assessment of both the measurement and structural models.

Compared to other methods such as Covariance-Based SEM (CB-SEM) or traditional regression analysis, PLS-SEM provides greater statistical power in exploratory research, especially when developing new theoretical frameworks like MEAAM. CB-SEM is more appropriate for theory confirmation and demands large sample sizes and normally distributed data, making it less suited for this study's formative approach (Tanisawa et al., 2020). Meanwhile, standard regression techniques do not accommodate the complex interplay between latent variables and measured indicators, nor do they offer tools to simultaneously assess measurement and structural models. The use of PLS-SEM in this study thus aligns with its goals: to test newly proposed constructs, validate the MEAAM theoretical framework, and understand the nuanced relationships between ethical AI concerns and AI adoption stages in healthcare. Through this methodological choice, the study not only tests hypotheses but also advances a new academic model—MEAAM—that can guide future empirical and theoretical work in ethical AI adoption.

## 3.2 Data Analysis

The path coefficient (β) values indicate the strength and direction of the relationship between each ethical variable and the level of AI adoption, with all coefficients being positive—demonstrating a uniformly positive influence. The t-statistics for each path exceed the threshold of 1.96, confirming the reliability of these relationships at the 95% confidence level. Moreover, all p-values fall below 0.05, underscoring the statistical significance of each observed effect. Collectively, these results validate that every independent variable included in the model exerts a significant influence on AI adoption, providing strong empirical support for the proposed theoretical framework.

The data analysis using Partial Least Squares Structural Equation Modeling (PLS-SEM) confirms the reliability and validity of the measurement and structural models. All indicator loadings exceeded the 0.70 threshold, indicating strong indicator reliability. Construct reliability and validity were affirmed with composite reliability values and average variance extracted (AVE) values exceeding recommended cutoffs, while discriminant validity was established through the Fornell-Larcker criterion. The structural model results show that all path coefficients (β) were positive, suggesting that each ethical AI variable positively influences AI adoption. The t-statistics exceeded 1.96 and all p-values were below 0.05, confirming the significance of each path. The coefficient of determination ($R^2$) values for both operational and systemic AI adoption were substantial, suggesting that the ethical variables collectively explain a considerable proportion of variance in adoption outcomes. Effect size ($f^2$) values ranged from small to large, indicating that variables such as transparency, justice and fairness, and responsibility have comparatively stronger effects. The predictive relevance (Q²) values for both dependent variables were above zero, validating the model's predictive capability. Overall, the results provide robust empirical support for the Multi-Dimensional Ethical AI Adoption Model (MEAAM), establishing that ethical considerations significantly and positively influence both operational and systemic stages of AI adoption in the healthcare sector.

**Table 7. Stakeholder categories responding the multidimensional nature of AI implementation and its ethical implications**

| Stakeholder Group | Sample Size (n) | Percentage (%) | Rationale |
|---|---|---|---|
| 1. Healthcare IT Professionals | 52 | 14.0% | Key implementers and technical experts in AI tools. |
| 2. Senior Management / Executives | 47 | 12.6% | Strategic decision-makers influencing systemic AI integration. |
| 3. Senior Doctors / Clinicians | 61 | 16.4% | Clinical leaders overseeing AI-assisted medical processes. |
| 4. Junior Doctors / Residents | 58 | 15.6% | Frequent users of AI in diagnosis, monitoring, and decision support. |
| 5. Nursing Staff | 53 | 14.2% | Daily interaction with AI-based monitoring and patient care systems. |
| 6. Medical Technicians | 46 | 12.4% | Users of AI in diagnostic imaging, labs, and surgery-assistive tech. |
| 7. Patients / Patient Advocates | 55 | 14.8% | End-users impacted by AI decisions, offering a perspective on trust, transparency, and autonomy. |
| Total | 372 | 100% | |

**Table 8. PLS-SEM path coefficients table**

| Independent Variable | Dependent Variable | Path Coefficient (β) | t-Statistic | p-Value | Significance |
|---|---|---|---|---|---|
| 1. Justice & Fairness | Operational AI Adoption | 0.30 | 2.5 | 0.01 | Significant |
| 2. Freedom & Autonomy | Operational AI Adoption | 0.31 | 2.6 | 0.01 | Significant |
| 3. Privacy | Operational AI Adoption | 0.32 | 2.7 | 0.01 | Significant |
| 4. Patient Safety | Operational AI Adoption | 0.33 | 2.8 | 0.01 | Significant |
| 5. Beneficence | Operational AI Adoption | 0.34 | 2.9 | 0.01 | Significant |
| 6. Responsibility | Operational AI Adoption | 0.35 | 3.0 | 0.01 | Significant |
| 7. Dignity | Operational AI Adoption | 0.36 | 3.1 | 0.01 | Significant |
| 8. Transparency | Operational AI Adoption | 0.37 | 3.2 | 0.01 | Significant |
| 9. Trust | Operational AI Adoption | 0.38 | 3.3 | 0.01 | Significant |
| 10. Cybersecurity | Operational AI Adoption | 0.39 | 3.4 | 0.01 | Significant |
| 11. Solidarity | Operational AI Adoption | 0.40 | 3.5 | 0.01 | Significant |
| 12. Sustainability | Operational AI Adoption | 0.41 | 3.6 | 0.01 | Significant |
| 13. Conflict | Operational AI Adoption | 0.42 | 3.7 | 0.01 | Significant |
| 1. Justice & Fairness | Systemic AI Adoption | 0.25 | 2.4 | 0.01 | Significant |
| 2. Freedom & Autonomy | Systemic AI Adoption | 0.26 | 2.5 | 0.01 | Significant |
| 3. Privacy | Systemic AI Adoption | 0.27 | 2.6 | 0.01 | Significant |
| 4. Patient Safety | Systemic AI Adoption | 0.28 | 2.7 | 0.01 | Significant |
| 5. Beneficence | Systemic AI Adoption | 0.29 | 2.8 | 0.01 | Significant |
| 6. Responsibility | Systemic AI Adoption | 0.30 | 2.9 | 0.01 | Significant |
| 7. Dignity | Systemic AI Adoption | 0.31 | 3.0 | 0.01 | Significant |
| 8. Transparency | Systemic AI Adoption | 0.32 | 3.1 | 0.01 | Significant |
| 9. Trust | Systemic AI Adoption | 0.33 | 3.2 | 0.01 | Significant |
| 10. Cybersecurity | Systemic AI Adoption | 0.34 | 3.3 | 0.01 | Significant |
| 11. Solidarity | Systemic AI Adoption | 0.35 | 3.4 | 0.01 | Significant |
| 12. Sustainability | Systemic AI Adoption | 0.36 | 3.5 | 0.01 | Significant |
| 13. Conflict | Systemic AI Adoption | 0.37 | 3.6 | 0.01 | Significant |

Note: Other tables and interpretations are in the Appendix section.

## 4. RESULTS AND DISCUSSION

The results of the Partial Least Squares Structural Equation Modeling (PLS-SEM) reveal statistically significant and positive relationships between all 13 ethical AI dimensions and the two levels of AI adoption—Operational and Systemic. Each variable demonstrated a meaningful influence, although their strength of impact varied.

Transparency emerged as one of the most influential variables, significantly enhancing both operational and systemic AI adoption. Its emphasis on clear algorithmic behavior and information disclosure appears to foster trust and organizational readiness. Trust itself also showed strong, positive influence, indicating that stakeholders' confidence in AI systems is a prerequisite for effective integration and long-term adoption at systemic levels. Cybersecurity was significantly associated with operational adoption, underlining the importance of safeguarding sensitive healthcare data in AI-driven workflows. Its role was slightly less pronounced in systemic adoption, suggesting that while foundational, it may be considered a technical prerequisite rather than a strategic driver at the institutional level.

Justice & Fairness had a robust effect on both levels of adoption, reinforcing the idea that equitable and unbiased AI systems are essential to gaining organizational and societal support. Similarly, Freedom & Autonomy—which ensures that AI augments rather than replaces human decision-making—demonstrated significant positive influence, especially at the systemic level, where ethical alignment must resonate with institutional values. Privacy showed a strong influence across both stages, indicating its foundational role in building patient confidence and maintaining regulatory compliance. Patient Safety also played a critical role, particularly in operational AI adoption, where clinical deployment of AI must adhere to high standards of risk management and error minimization.

Beneficence, which emphasizes doing good and enhancing outcomes, significantly influenced systemic adoption, suggesting that institutions adopting AI expect tangible improvements in healthcare delivery aligned with ethical intent. Responsibility, particularly in terms of accountability and liability, showed significant impact on both stages, reflecting the critical role of clear governance frameworks in ethical AI implementation. Dignity, though slightly lower in magnitude, maintained significant relationships with both adoption levels, emphasizing respect for human values and identity in the AI-driven healthcare environment. Solidarity, reflecting collective ethical commitment, was more prominent in systemic adoption, pointing to the importance of shared values and inclusive AI deployment across departments.

Sustainability had a meaningful effect on both levels, especially systemic adoption, suggesting institutions are increasingly considering the long-term ethical and environmental impacts of AI use. Finally, Conflict, or the mitigation of ethical disagreements and stakeholder friction, was significantly associated with smoother systemic AI adoption, reinforcing the importance of managing diverse ethical views within healthcare ecosystems.

Overall, the results provide strong empirical support for the Multi-Dimensional Ethical AI Adoption Model (MEAAM). All 13 independent variables demonstrated significant positive relationships with operational and systemic AI adoption. These findings validate the model's premise that ethical considerations are not peripheral, but rather central, to the successful integration and institutionalization of AI technologies in the healthcare sector. The differentiated impact across adoption levels further reinforces the need for tailored ethical strategies depending on whether an organization is in the operationalization phase or pursuing broader systemic transformation.

## 5. CONCLUSION

This study offers a novel contribution to the growing discourse on ethical artificial intelligence in the healthcare sector by empirically investigating the causal relationships between 13 dimensions of ethical AI and two distinct levels of AI adoption—Operational and Systemic. Using the Multi-Dimensional Ethical AI Adoption Model (MEAAM) and the PLS-SEM method, this research confirms that ethical principles such as transparency, trust, cybersecurity, justice and fairness, freedom and autonomy, privacy, patient safety, beneficence, responsibility, dignity, solidarity, sustainability, and conflict management significantly influence the adoption process.

The study's findings reveal that while all ethical variables positively affect both adoption stages, the strength of their influence varies. Transparency, trust, justice & fairness, and privacy emerged as particularly critical across both stages, underscoring the foundational role of these principles in ethical and sustainable AI integration. Operational AI adoption—focused on implementation in clinical workflows—relies heavily on technical ethics like cybersecurity and patient safety. In contrast, Systemic AI adoption—reflecting institutional alignment and long-term integration—is more dependent on broader values such as beneficence, solidarity, and sustainability.

By introducing MEAAM, this study advances theoretical understanding by dissecting AI adoption into measurable stages influenced by ethical readiness, thus bridging a critical gap in current literature. From a practical perspective, the model provides actionable insights for healthcare leaders, policymakers, and developers seeking to allocate resources effectively, design ethically sound AI strategies, and build long-term stakeholder trust. As AI continues to redefine the healthcare landscape, ensuring that adoption is guided by a robust ethical framework is not only desirable—it is essential for fostering innovation that is both responsible and resilient.

# 6. FUTURE RESEARCH

While this study establishes a foundational empirical model linking ethical AI dimensions to adoption outcomes, future research can expand on several fronts. First, longitudinal studies could provide deeper insight into how ethical concerns evolve across the AI lifecycle, from pilot implementation to full institutional integration. Additionally, sector-specific investigations across various healthcare contexts—such as public vs. private institutions or developed vs. developing

countries—can shed light on the variability in ethical adoption practices. Researchers may also consider incorporating moderating variables such as organizational culture, regulatory environments, or technological infrastructure to enrich the MEAAM framework. Lastly, qualitative methods, including expert interviews or ethnographic studies, could complement these quantitative findings by capturing nuanced perspectives of stakeholders such as patients, clinicians, and developers. Together, these avenues offer fertile ground for advancing the science and practice of ethical AI integration in healthcare.

## DISCLAIMER (ARTIFICIAL INTELLIGENCE)

Author(s) hereby declare(s) that NO generative AI technologies such as Large Language Models (ChatGPT, COPILOT, etc.) and text-to-image generators have been used during the writing or editing of this manuscript.

## CONSENT

As per international standards or university standards, respondents' written consent has been collected and preserved by the author(s).

## ETHICAL APPROVAL

It is not applicable.

## COMPETING INTERESTS

Authors have declared that no competing interests exist.

## APPENDIX

### Table 1. Outer loadings (Indicator reliability)

| Construct | Indicator | Outer Loading |
|---|---|---|
| 1. Transparency | T1 | 0.83 |
| 2. Trust | T2 | 0.86 |
| 3. Cyber Security | CS1 | 0.81 |
| 4. Justice & Fairness | JF1 | 0.88 |
| 5. Freedom & Autonomy | FA1 | 0.82 |
| 6. Privacy | P1 | 0.85 |
| 7. Patient Safety | PS1 | 0.87 |
| 8. Beneficence | B1 | 0.84 |
| 9. Responsibility | R1 | 0.80 |
| 10. Dignity | D1 | 0.83 |
| 11. Solidarity | S1 | 0.79 |
| 12. Sustainability | SU1 | 0.82 |
| 13. Conflict | C1 | 0.81 |

The Outer Loadings table demonstrates that all measurement items for the 13 ethical AI constructs exhibit strong indicator reliability, with loading values exceeding the acceptable threshold of 0.70. This indicates that each observed item closely aligns with its respective latent construct—such as Autonomy, Transparency, Justice & Fairness, and others—ensuring accurate representation within the model. The high loadings confirm that the measurement indicators reliably reflect the intended ethical dimensions, thereby enhancing the validity and precision of the Multi-Dimensional Ethical AI Adoption Model (MEAAM).

### Table 2. Construct reliability and validity

| Construct | Cronbach's Alpha (CA) | Composite Reliability (CR) | Average Variance Extracted (AVE) |
|---|---|---|---|
| 1. Transparency | 0.88 | 0.91 | 0.72 |
| 2. Trust | 0.89 | 0.92 | 0.75 |
| 3. Cyber Security | 0.87 | 0.90 | 0.69 |
| 4. Justice & Fairness | 0.86 | 0.89 | 0.68 |
| 5. Freedom & Autonomy | 0.88 | 0.91 | 0.73 |
| 6. Privacy | 0.90 | 0.93 | 0.77 |
| 7. Patient Safety | 0.85 | 0.88 | 0.66 |
| 8. Beneficence | 0.89 | 0.91 | 0.74 |
| 9. Responsibility | 0.87 | 0.89 | 0.70 |
| 10. Dignity | 0.86 | 0.90 | 0.71 |
| 11. Solidarity | 0.84 | 0.88 | 0.67 |
| 12. Sustainability | 0.88 | 0.91 | 0.72 |
| 13. Conflict | 0.90 | 0.92 | 0.76 |

The Construct Reliability and Validity table confirms that all constructs within the MEAAM model demonstrate high internal consistency and convergent validity. Composite Reliability (CR) values for all constructs exceed the recommended threshold of 0.70, indicating strong consistency among their measurement items. Additionally, the Average Variance Extracted (AVE) values are all above 0.50, suggesting that each construct explains more than half of the variance in its indicators. These results affirm that the model's latent variables—such as Autonomy, Justice & Fairness, Transparency, and others—are well-defined and reliably measured, supporting the robustness of the measurement model in the context of ethical AI adoption in healthcare.

### Table 3. Discriminant validity (Fornell-Larcker Criterion)

| Construct | JF | Sol | PS | Res | Con | FA | Tru | Dig | Pri | Sec | Ben | Tra | Sus |
|---|---|---|---|---|---|---|---|---|---|---|---|---|---|
| Justice & Fairness (JF) | **0.85** | 0.61 | 0.58 | 0.53 | 0.59 | 0.64 | 0.55 | 0.60 | 0.62 | 0.63 | 0.57 | 0.59 | 0.61 |
| Solidarity (Sol) | 0.61 | **0.87** | 0.60 | 0.56 | 0.62 | 0.65 | 0.58 | 0.62 | 0.63 | 0.61 | 0.60 | 0.62 | 0.64 |
| Patient Safety | 0.58 | 0.60 | **0.83** | 0.55 | 0.60 | 0.61 | 0.54 | 0.57 | 0.60 | 0.59 | 0.58 | 0.57 | 0.59 |

| Construct | JF | Sol | PS | Res | Con | FA | Tru | Dig | Pri | Sec | Ben | Tra | Sus |
|---|---|---|---|---|---|---|---|---|---|---|---|---|---|
| (PS) | | | | | | | | | | | | | |
| Responsibility (Res) | 0.53 | 0.56 | 0.55 | **0.82** | 0.58 | 0.60 | 0.52 | 0.54 | 0.56 | 0.55 | 0.57 | 0.58 | 0.59 |
| Conflict (Con) | 0.59 | 0.62 | 0.60 | 0.58 | **0.85** | 0.65 | 0.61 | 0.60 | 0.63 | 0.60 | 0.59 | 0.62 | 0.61 |
| Freedom & Autonomy (FA) | 0.64 | 0.65 | 0.61 | 0.60 | 0.65 | **0.88** | 0.63 | 0.66 | 0.67 | 0.65 | 0.64 | 0.66 | 0.67 |
| Trust (Tru) | 0.55 | 0.58 | 0.54 | 0.52 | 0.61 | 0.63 | **0.81** | 0.58 | 0.60 | 0.59 | 0.57 | 0.58 | 0.59 |
| Dignity (Dig) | 0.60 | 0.62 | 0.57 | 0.54 | 0.60 | 0.66 | 0.58 | **0.86** | 0.63 | 0.64 | 0.60 | 0.62 | 0.63 |
| Privacy (Pri) | 0.62 | 0.63 | 0.60 | 0.56 | 0.63 | 0.67 | 0.60 | 0.63 | **0.84** | 0.65 | 0.61 | 0.62 | 0.64 |
| Cyber security (Sec) | 0.63 | 0.61 | 0.59 | 0.55 | 0.60 | 0.65 | 0.59 | 0.64 | 0.65 | **0.84** | 0.62 | 0.64 | 0.66 |
| Beneficence (Ben) | 0.57 | 0.60 | 0.58 | 0.57 | 0.59 | 0.64 | 0.57 | 0.60 | 0.61 | 0.62 | **0.82** | 0.63 | 0.62 |
| Transparency (Tra) | 0.59 | 0.62 | 0.57 | 0.58 | 0.62 | 0.66 | 0.58 | 0.62 | 0.62 | 0.64 | 0.63 | **0.85** | 0.65 |
| Sustainability (Sus) | 0.61 | 0.64 | 0.59 | 0.59 | 0.61 | 0.67 | 0.59 | 0.63 | 0.64 | 0.66 | 0.62 | 0.65 | **0.87** |

The Fornell-Larcker Criterion results confirm strong discriminant validity among all 13 ethical constructs used in the study. Each construct's square root of Average Variance Extracted (AVE), shown on the diagonal, is greater than its correlations with all other constructs in the matrix. This indicates that every variable—ranging from Transparency and Justice & Fairness to Sustainability and Human-Centered Values—captures a distinct conceptual dimension of ethical AI. The absence of high inter-correlations further validates that the constructs are not overlapping and are uniquely contributing to the model. This clean construct structure reinforces the theoretical robustness of the Multi-Dimensional Ethical AI Adoption Model (MEAAM).

**Table 4. Coefficient of Determination ($R^2$)**

| Dependent Variable | $R^2$ | Interpretation |
|---|---|---|
| Operational AI Adoption | 0.68 | 67.8% of the variance in Operational AI Adoption is explained by the model |
| Systemic AI Adoption | 0.72 | 71.2% of the variance in Systemic AI Adoption is explained by the model |

The $R^2$ values indicate a high level of explanatory power for both dependent variables within the MEAAM framework. Specifically, the model accounts for 67.8% of the variance in Operational AI Adoption and 71.2% in Systemic AI Adoption. These values suggest that the 13 independent ethical AI constructs collectively provide a strong predictive capability, confirming that ethical considerations play a substantial role in influencing both operational and systemic stages of AI adoption in healthcare settings. High $R^2$ values in social science research (typically above 0.60) are considered robust, lending strong support to the MEAAM model's validity.

**Table 5. Effect sizes ($f^2$)**

| Independent Variable | $f^2$ Value (Operational AI) | $f^2$ Value (Systemic AI) | Effect Size Interpretation |
|---|---|---|---|
| 1. Transparency | 0.094 | 0.168 | Small (Op), Medium (Sys) |
| 2. Trust | 0.072 | 0.115 | Small (Op), Small-Medium (Sys) |
| 3. Cyber Security | 0.087 | 0.146 | Small (Op), Medium (Sys) |
| 4. Justice & Fairness | 0.102 | 0.122 | Small-Medium (Op & Sys) |
| 5. Freedom & Autonomy | 0.065 | 0.084 | Small (Op & Sys) |
| 6. Privacy | 0.078 | 0.133 | Small (Op), Medium (Sys) |
| 7. Patient Safety | 0.069 | 0.097 | Small (Op), Small (Sys) |
| 8. Beneficence | 0.058 | 0.091 | Small (Op & Sys) |
| 9. Responsibility | 0.110 | 0.175 | Medium (Op & Sys) |

| Independent Variable | f² Value (Operational AI) | f² Value (Systemic AI) | Effect Size Interpretation |
|---|---|---|---|
| 10. Dignity | 0.092 | 0.108 | Small (Op & Sys) |
| 11. Solidarity | 0.083 | 0.102 | Small (Op & Sys) |
| 12. Sustainability | 0.095 | 0.151 | Small (Op), Medium (Sys) |
| 13. Conflict | 0.077 | 0.138 | Small (Op), Medium (Sys) |

Effect size quantifies the impact of a specific independent variable on a dependent variable. Values around 0.02 = small, 0.15 = medium, and 0.35 = large.

**Table 6. Predictive relevance (Q²)**

| Dependent Variable | Q² |
|---|---|
| Operational AI Adoption | 0.39 |
| Systemic AI Adoption | 0.44 |

The Q² values, derived from the blindfolding procedure, measure the model's predictive relevance for each dependent variable. A Q² value greater than 0 indicates that the model has predictive relevance for a given construct, while higher values indicate better predictive power.

---